\documentclass[final,5p,twocolumn]{elsarticle}

\usepackage{amssymb}
\usepackage{xcolor}
\usepackage{graphicx}
\usepackage{multirow}
\usepackage{tabularx} % For tables with equal-width columns
\usepackage{booktabs} % For prettier tables
\graphicspath{ {images/} }

\journal{Neurocomputing}

\begin{document}

\begin{frontmatter}



%\title{}


%\author{}

%\address{}

%\title{Logo Recognition 2 - La vendetta}
\title{Deep Learning for Logo Recognition}
\author{Simone Bianco, Marco Buzzelli, Davide Mazzini, Raimondo Schettini}
%\address{DISCo (Dipartimento di Informatica, Sistemistica e Comunicazione),
%Universit\`{a} degli Studi di Milano-Bicocca, viale Sarca 336, 20126 Milano, Italy}
%%\\ \email{simone.bianco@disco.unimib.it, marco.buzzelli@disco.unimib.it, davide.mazzini@disco.unimib.it, schettini@disco.unimib.it}
%%}

\address{DISCo - Universit\`{a} degli Studi di Milano-Bicocca, 20126 Milano, Italy}
%\\ \email{simone.bianco@disco.unimib.it, marco.buzzelli@disco.unimib.it, davide.mazzini@disco.unimib.it, schettini@disco.unimib.it}
%}

\begin{abstract}
%{\color{black!40!orange}
In this paper we propose a method for logo recognition using deep learning. Our recognition pipeline is composed of a logo region proposal followed by a Convolutional Neural Network (CNN) specifically trained for logo classification, even if they are not precisely localized. %by the region proposal algorithm.
%explicitly taking into account imprecise region proposed
%State of the art approaches use instead networks pretrained on a different task, due to the lack of enough publicly-available training data.
Experiments are carried out on the FlickrLogos-32 database, and we evaluate the effect on recognition performance of synthetic versus real data augmentation, and image pre-processing. Moreover, we systematically investigate the benefits of different training choices such as class-balancing, sample-weighting and explicit modeling the background class (i.e. no-logo regions). Experimental results confirm the feasibility of the proposed method, that outperforms the methods in the state of the art.
%}
\end{abstract}


\end{frontmatter}


\section{Introduction}
\label{sec:intro}
%{\color{black!40!green}Logo recognition is ...}
%{\color{black!40!green}
Logo recognition in images and videos is the key problem in a wide range of applications, such as copyright infringement detection, contextual advertise placement, vehicle logo for intelligent traffic-control systems  \cite{psyllos2010vehicle}, automated computation of brand-related statistics on social media \cite{gao2014brand}, augmented reality \cite{hagbi2011shape}, etc.
%}
%Such problems are typically addressed with keypoint-based detectors and descriptors like SIFT \cite{romberg2013bundle,bianco2013quantitative,bianco2015local}. These methods are in fact best suited for well-defined shapes and affine transformations, like those found in the domain of logos. Since in many real applications the logo images could be highly degradated, in this paper we investigated Convolutional Neural Networks \cite{fukushima1980neocognitron} as an alternative approach that is not based on keypoint detection.
%CNNs fall in the category of Deep Learning techniques, which have been employed in others fields such as speech recognition \cite{abdel2014convolutional} and action recognition \cite{foggia2014exploiting}.}\\

%{\color{black!40!green}
Traditionally, logo recognition has been addressed with keypoint-based detectors and descriptors \cite{bagdanov2007trademark,kleban2008spatial,joly2009logo,meng2010interactive}. For example Romberg and Lienhart \cite{romberg2013bundle} presented a scalable logo recognition technique based on feature bundling, where individual local features are aggregated with features from their spatial neighborhood into Bag of Words (BoW). Romberg et al. \cite{romberg2011scalable} exploited a method for encoding and indexing the relative spatial layout of local features detected in the logo images. Based on the analysis of the local features and the composition of basic spatial structures, such as edges and triangles, they derived a quantized representation of the regions in the logos. Revaud et al. \cite{revaud2012correlation} introduced a technique to down-weight the score of those noisy logo detections by learning a dedicated burstiness model for the input logo. {Boia et al. \cite{boia2016logo,boia2015elliptical} proposed a smart method to perform both logo localization and recognition using homographic class graphs. They also exploited inverted secondary models to handle inverted colors instances.}
Recently some works investigating the use of deep learning for logo recognition appeared \cite{bianco2015logo,eggert2015benefit,iandola2015deeplogo}.
%In our previous work \cite{bianco2015logo} we started investigating the use of deep learning for logo recognition,
%%on the FlickrLogos32 dataset \cite{romberg2011scalable},
%where we tried different techniques to deal with the limited amount of training data. We exploited a pretrained network and performed an augmentation of the training set with artificial transformations. The proposed recognition pipeline achieved state of the art results on the FlickrLogos32 dataset \cite{romberg2011scalable} and demonstrated to be robust to blur, noise and lossy compression.
%
Bianco et al. \cite{bianco2015logo} and  Eggert et al. \cite{eggert2015benefit} investigated the use of pretrained Convolutional Neural Networks (CNN) and synthetically generated data for logo recognition,
%on the FlickrLogos32 dataset \cite{romberg2011scalable},
trying different techniques to deal with the limited amount of training data. %They exploited a pretrained network and performed an augmentation of the training set with artificial transformations. %The proposed recognition pipeline achieved state of the art results on the FlickrLogos32 dataset \cite{romberg2011scalable} and demonstrated to be robust to blur, noise and lossy compression.
%
%A similar approach was investigated by Eggert et al. \cite{eggert2015benefit}
%to explore the benefits of synthetically generated data for the task of company logo detection with
%deep-learned features in the absence of a large training set.
Also Iandola et al. \cite{iandola2015deeplogo} investigated a similar approach, proposing and evaluating several network architectures. {Oliveira et al. \cite{oliveira2016automatic} exploited pretrained CNN models and used them as part of a Fast Region-Based Convolutional Networks recognition pipeline.}
Given the limited amount of training data available for the logo recognition task, all these methods work on networks pretrained on different tasks.%}

%In this paper we extend the investigation of deep learning for logo recognition using a pipeline composed by a candidate logo region proposal followed by a Convolutional Neural Network for logo classification.
In this paper we propose a method for logo recognition exploiting deep learning. The recognition pipeline is composed by a recall-oriented logo region proposal \cite{girshick2016region}, followed by a Convolutional Neural Network (CNN) specifically trained for logo classification, even if they are not precisely localized.
Within this pipeline, we investigate the benefit on the recognition performance of the application of different machine learning techniques in training, such as image pre-processing, class-balancing, sample weighting, and synthetic data augmentation.
Furthermore we prove the benefit of adding as positive examples candidate regions coming from the object proposal to the ground truth logos, and the benefit of enlarging the size of the actual dataset with real data augmentation and the use of a background class (i.e. no-logo regions) in training.

%using actual investigate t
%in order to be able to train a deep neural network from scratch instead of using a pretrained one, we introduce an expanded dataset for logo recognition on the 32 classes defined in FlickrLogos-32 \cite{romberg2011scalable}.
%Such new dataset is used to train a deep neural network.

%benefits and crosstalks between also perform several tests to assess the contribution to the overall results given by the application of

%\begin{figure}
%	\includegraphics[width=\textwidth]{pipeline.png}
%	\centering
%	\caption{[...]}
%	\label{fig:pipeline}
%\end{figure}

%%%%%%%%%%%%%%%%%%%%%%%%%%%%%%%%%%%%%%%%%%%%%%%%%%%%%%%%%%%%%%%
\section{Proposed Method}
\label{sec:proposedmethod}
\begin{figure*}[htbp]
	\includegraphics[width=0.9\textwidth]{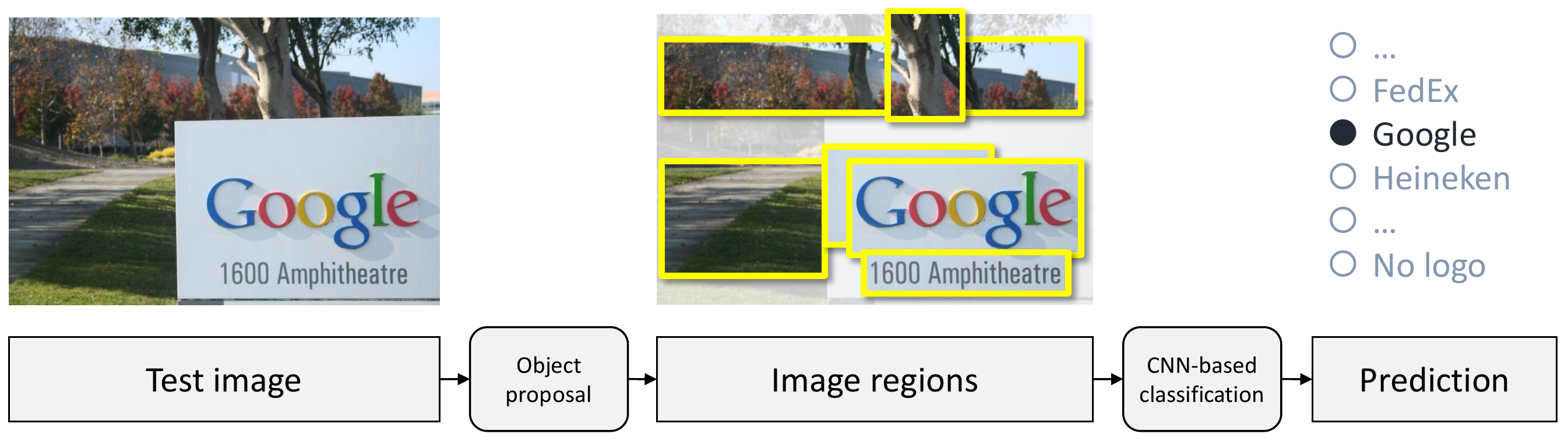}
	\centering
	\caption{Simplified logo classification pipeline}
	\label{fig:pipeline_simplified}
\end{figure*}

%In this section we describe the pipelines used for training and testing.
%{\color{black!40!green}
%\textcolor{red}{All'interno di un framework dove ci sono un coso proposal e una rete cnn che fa roba, ci siamo chiesti come addestrare una rete che fosse ...}
%{\color{black!40!green}
The proposed classification pipeline is illustrated in Figure \ref{fig:pipeline_simplified}.
%For our experiments we use the classification pipeline reported in Figure \ref{fig:pipeline_simplified}, which has been inspired from the object recognition domain \cite{girshick2016region} and has been used in different variants in \cite{bianco2015logo,eggert2015benefit,iandola2015deeplogo}:
Since logos may appear in any image location with any orientation and scale, and more logos can coexist in the same image, for each image we generate different object proposals, that are regions which are more likely to contain a logo.
%The algorithm used for the extraction of the objects proposals is class-agnostic, therefore it extracts regions of different aspect-ratios that can be used to recognize objects under different kinds of geometric transformation.
These proposal are then cropped to a common size to match the input dimensions of the neural network and are propagated through a CNN specifically trained for logo recognition.
%Images are limited to a maximum side length of 1024 pixels and are rescaled if necessary.
%Each propoposal defines
%an image patch which is resized to match the input dimensions
%of the neural network and is propagated through the
%DCNN. For our experiments, we use the 16-layer very deep
%architecture by [5].\\
%...\\
%In order to have an effective classification and recognition using a CNN approach, we have designed a novel training procedure that takes into account not only the data variablity within the images to be processeeed but also the localization errror of the object proposal.\\
%...\\
%Within the operational framework depicted in Figure \ref{fig:pipeline}, which is composed by an object proposal module followed by a CNN for logo recognition and classification, we asked ourselves how to train a CNN robust to the different transformation that a logo can undergo in typical images and to bad object proposals. In fact, given an input image, the framework extracts
%regions which are more likely to contain an object. These regions are called object
%proposals. The algorithm used for the extraction of the objects proposals is class-
%agnostic, therefore it extracts regions of different aspect-ratios that can be used
%to recognize objects under different kinds of geometric transformation. These
%proposal are then warped to a common size and processed for
%query expansion in order to increase recall. Finally we use a pre-trained CNN as
%feature extractor and a linear SVM for logo recognition and classification. Figure
%1 shows the main steps of the recognition pipeline.}

In order to have performance as high as possible within this pipeline, we use an object proposal that is highly recall-oriented.  For this reason, the CNN classifier should be designed and trained to take into account that the logo regions proposed may contain many false positives or only parts of actual logos. To address these problems we propose here a training framework and investigate the influence on the final recognition performance of different implementation choices.

\begin{figure*}[htbpp]
	\includegraphics[width=0.9\textwidth]{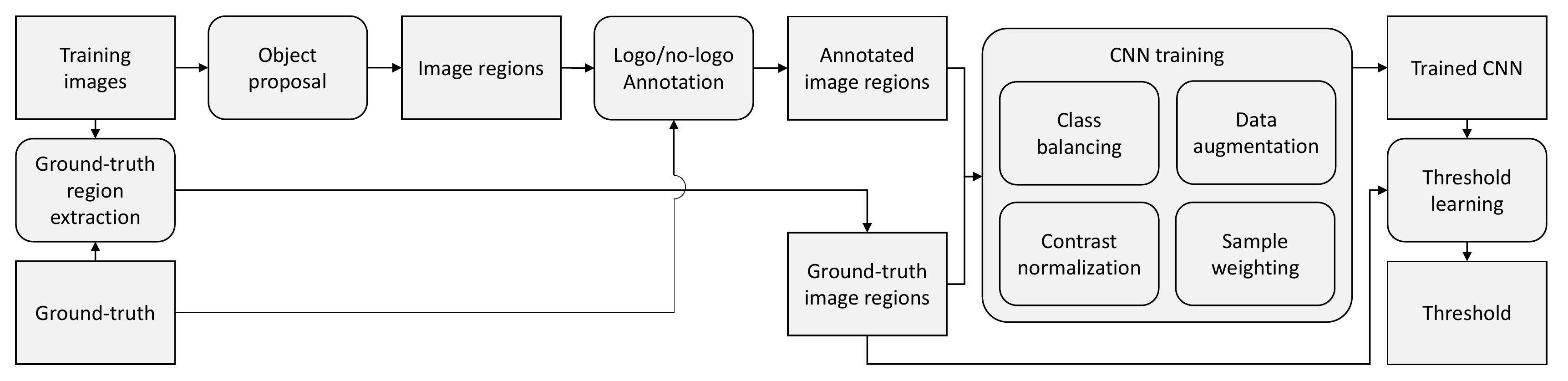}
	\centering
	\caption{Logo recognition training framework.}
	\label{fig:pipe_training}
\end{figure*}

In more detail, the
%}
%A schematic representation of the
training framework is reported in Figure \ref{fig:pipe_training}. The training data preparation is composed by two main parts:
\begin{itemize}
\item[-] {\bf Precise ground-truth logo annotations}: Given a set of training images and associated ground-truth specifying logo position and class, we first crop logo regions and annotate them with the ground-truth class. These regions are rectangular crops that completely contain logos but, due to the prospective of the image or the logo particular shape, may also contain part of the background.% Some example of cropped logos are shown in \textcolor{red}{Figure \ref{fig:XXX}}.
\item[-] {\bf Object-proposal logo annotations}: Since we must automatically localize regions that may contain a logo, an object proposal algorithm is employed in the whole pipeline as shown in Figure \ref{fig:pipeline_simplified}. This algorithm is not applied only to the test images, but it is also run on the training images to extract regions that are more likely to contain a logo. Details about the particular algorithm used are given in the next subsection.
%{\color{black!40!orange}
Each object proposal in the training images is then labeled on the basis of its content: if it overlaps with a ground-truth logo region, it is annotated with the corresponding class and with the Intersection-over-Union (IoU) overlap ratio, otherwise it is labeld as background.%}
\end{itemize}

%Given a set of training images, we extract regions which are more likely to contain an object. These regions are called object proposals.
%The algorithm used for the extraction of the objects proposals, Selective Search, extracts regions of different aspect-ratios that can be used to recognize objects under different kinds of geometric transformation.
Within our training framework we investigate both the use of the precise ground-truth logo annotations alone or coupled with the object-proposal logo annotations.
All positive instances, i.e. labeled logos and eventually object proposals that overlap with them by a significant amount (i.e.  IoU$\geq 0.5$), are used to train a Convolutional Neural Network whose architecture is given below.
Different training choices are investigated within our framework in Figure \ref{fig:pipe_training}: %Each contribution to the performance of the proposed method will be evaluated in Section \ref{sec:results}.

%{\color{black!40!green}
\begin{itemize}
\item[-] {\bf{Class balancing}}: The logo classes are balanced by replicating the examples of classes with lower cardinality. Two different strategies are implemented: epoch-balancing, where classes are balanced in each training epoch, and batch-balancing, where classes are balanced in each training batch. The hypothesis is that this should prevent a classification bias of the CNN.
\item[-] {\bf{Data augmentation}}: Training examples are augmented in number by generating random shifts of logo regions. The hypothesis is that this should make the CNN more robust to inaccurate logo localization at test time.
\item[-] {\bf{Contrast normalization}}: Images are contrast-normalized by subtracting the mean and dividing by the standard deviation, which are extracted from the whole training set. The hypothesis is that this should make the CNN more robust to changes in the lighting and imaging conditions.
\item[-] {\bf{Sample weighting}}: Positive instances are weighted on the basis of their overlap with ground-truth logo regions. The hypothesis is that this should make the CNN more confident on proposals highly overlapping with the ground truth logos.
\item[-] {\bf{Background class}}: A background class is considered together with the logo classes.
%This contains background examples that do not overlap with any ground-truth region of the logo classes.
%{\color{black!40!orange}
Background examples are not randomly selected, but are composed by the candidate regions generated by the object proposal algorithm on training images and that do not overlap with any logo. %}
The hypothesis is that this should make the CNN more precise in discriminating logos and background class. %by explicitly modeling the backround class.
%\item[-] {\bf{Positive examples}}: Two different types of positive examples are considered. In the former only ground-truth logo region are used as positive examples for training; in the latter, object proposals with high overlap with ground-truth logo regions are also considered.
\end{itemize}
The actual contribution to the performance of each training choice considered will be discussed in Section \ref{sec:results}.

After the CNN is trained, a threshold is learned on top of the CNN predictions. If the CNN prediction with the highest confidence is below this threshold, the candidate region is labeled as not being a logo, otherwise CNN prediction is left unchanged.
%}
%Section \ref{sec:results} shows the contribution of each of these steps to the overall quality of the proposed method.
%Finally, we learn a threshold on a validation set, to be applied over the confidence score of each classification to accept it or label it as not being a logo.

\begin{figure*}[htbpp]
	\includegraphics[width=0.9\textwidth]{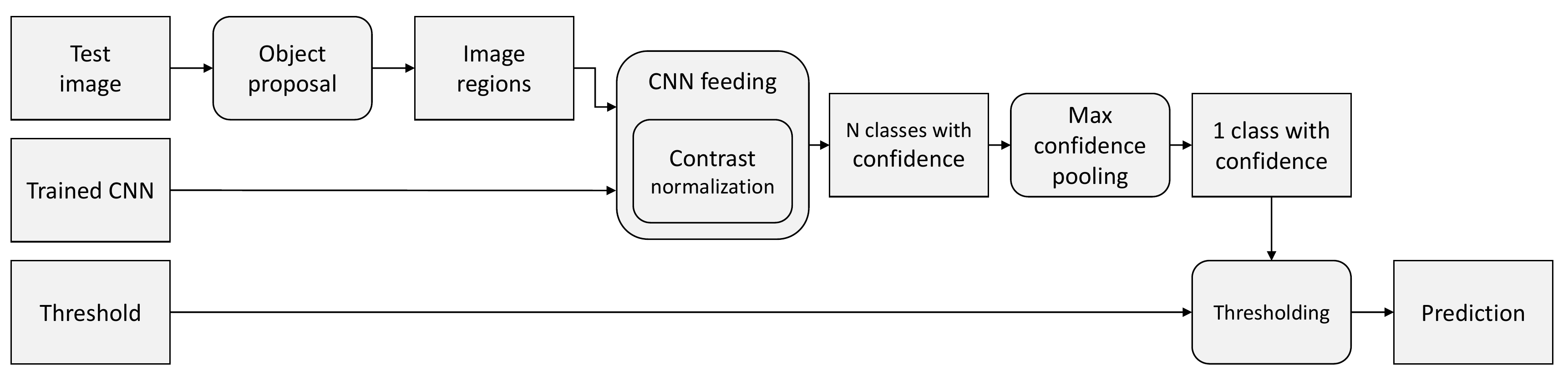}
	\centering
	\caption{Logo recognition testing framework.}
	\label{fig:pipe_test}
\end{figure*}

%{\color{black!40!green}
%A more detailed representation of
The testing framework is reported in Figure \ref{fig:pipe_test}.
Given a test image, we extract the object proposals with the same algorithm used for training.
We then perform contrast-normalization over each proposal (if enabled at training time), and feed them to the CNN. The CNN predictions on the proposals are max-pooled and the class identified with highest confidence (eventually including the background class) is selected. If the CNN confidence for a logo class is above the threshold that has been learned in training, the corresponding logo class is assigned to the image, otherwise the image is labeled as not containing any logo.%}

%Training:
%\begin{enumerate}
%	\item Object proposal with Selective Search
%	\item Data augmentation
%	\item Class-balancing
%	\item Preprocessing (Contrast normalization)
%	\item Neural Network training: 32 classes + 1 background class
%	\item Best threshold selection on validation set
%\end{enumerate}
%Testing:
%\begin{enumerate}
%	\item Object proposal with Selective Search
%	\item Preprocessing (Contrast normalization)
%	\item Neural Network testing
%	\item Max pooling for whole-image classification
%	\item Threshold to decide logo/nologo
%\end{enumerate}

\subsection{Object proposal}
\label{subsec:selective}
For object proposal we exploit a Selective Search algorithm originally introduced by van de Sande et al. \cite{vandeSandeICCV2011,UijlingsIJCV2013}.\\
The goal of Selective Search is to provide a set of regions likely to contain an instance of the object of interest, i.e. logos in our case. They can appear in any position and scale, and may have been acquired under different lighting conditions, and from slightly different point of views.
The algorithm is designed to be highly recall-oriented; this implies that very few logos are not segmented, but also implies that a great number of false positive candidates are generated. The proposed regions will be disambiguated by the neural network that comes afterward.
%Such approach makes it possible to find logos at different scales, acquired under unknown lighting conditions, and affected by perspective transformations.\\
% As an alternative solution, one might consider a multi-scale fully convolutional approach, producing an activation map that covers different positions and scales in the image. The limitation of this method is that, at larger scales, instances are badly located due the impact of stride from the network layers. On the other hand, Selective Search allows for a precise localization at all scales.

%Selective Search exploits a hierarchical grouping algorithm. Initial regions are created using superpixels segmentation \cite{felzenszwalb2004efficient}, and then a variety of complementary grouping criteria is used, such as: color similarity, texture similarity, proximity of small regions (encouraging small parts to merge early), and overlapping of regions (to merge objects composed of interleaving parts).
%These criteria are designed for fast computation: features are computed from image data only the first time, when analyizing the initial superpixels, and are subsequently propagated up the hierarchy by combining the features of merging regions.\\
%The final set of object proposals is then obtained by considering all levels of the generated hierarchy.\\
%Selective Search is designed to potentially exploit different color spaces. In this work we use the HSV representation, shown in \cite{bianco2015logo} to be the best solution for the task of logo recognition.

%%%%%%%%%%%%%%%%%%%%%%%%%%%%%%%%%%%%%%%%%%%%%%%%%%%%%%%%%%%%%%%
\subsection{Network Architecture}
\label{sec:architecture}
%The network architecture used for the experiments in the following sections is a tiny deep network structure used for the first time by Alex Krizhevsky in \cite{krizhevsky2012imagenet} on the CIFAR-10 dataset. It has three convolutional layers interleaved by ReLU nonlinearities and Pooling layers. All the pooling layers make the data dimensions halve after every Pooling block. The last part of the network (farthest from the input) consists in two Fully-connected layers with a final Softmax classifier. The whole net structure is presented in Table \ref{tab:net_architecture}.
The architecture used for the experiments in the following sections is a tiny deep neural network. We opted for a tiny network because it is fast at test time and it can be trained on cheap GPUs in very short time. It also allows us to train the network without using any form of regularization like dropout \cite{srivastava2014dropout}, dropconnect \cite{wan2013regularization}, etc. decreasing even more the time needed for training and validating the network.\\
%Table \ref{tab:timings} shows the timings of the whole recognition pipeline.
%The time needed to process all the patches extracted by the object proposal algorithm ($\sim$1000 per image) is 0.71 seconds on CPU and 0.36 on GPU. The training time on GPU takes about 20 minutes.\\
The same network structure was used by Krizhevsky in \cite{krizhevsky2012imagenet} on the CIFAR-10 dataset, where it was proven to be an high-performance network for the task of object recognition on tiny RGB images. It has three convolutional layers interleaved by ReLU nonlinearities and Pooling layers. All the pooling layers make the data dimensions halve after every Pooling block. The last part of the network (farthest from the input) consists in two Fully-connected layers with a final Softmax classifier. The whole net structure is presented in Table \ref{tab:net_architecture}.\\
%Results in Table \ref{tab:comparison} show that our small network structure, if trained with the proper configuration (see Table \ref{tab:results_thval}), achieves higher levels of accuracy with respect to huge networks used in \cite{eggert2015benefit} and \cite{iandola2015deeplogo}.
To give an idea of the network size, our network has $1.5 \times 10^5$ parameters whereas AlexNet (used in \cite{eggert2015benefit}) and GoogLeNet (a similar structure is used in \cite{iandola2015deeplogo}) have respectively $6 \times 10^7$ and $1.3 \times 10^7$ parameters. Therefore our network is less likely to overfit, even when the size of the training set is not large.

\begin{table}[htbpp]
	\caption{Neural Network Architecture}
	\label{tab:net_architecture}
	\center
\begin{tabular}{ll}
\toprule
 & \textbf{Layers}\\ \midrule
1 & Conv 32 filters of 5x5\\
2 & Pool (max) with stride 2\\
3 & Relu\\
4 & Conv 32 filters of 5x5\\
5 & Relu\\
6 & Pool (average) with stride 2\\
7 & Conv 64 filters of 5x5\\
8 & Relu\\
9 & Pool (average) with stride 2\\
10 & Fully Connected of size 64 \\
11 & Fully Connected of size 33 \\
12 & Softmax\\ \bottomrule
\end{tabular}
\end{table}

%The use of a tiny network makes it very fast at test time allowing to process all the patches extracted by the object proposal algorithm ($\sim$1000 per image) in 0.7 seconds on CPU (see also Table \ref{tab:timings}). It also allows us to train the network without using any form of regularization like dropout \cite{srivastava2014dropout}, dropconnect \cite{wan2013regularization}, etc.  decreasing the time needed for training and validating the network.

%%%%%%%%%%%%%%%%%%%%%%%%%%%%%%%%%%%%%%%%%%%%%%%%%%%%%%%%%%%%%%%
\section{Logos datasets}
%\section{FlickrLogos-32 Dataset}
\subsection{FlickrLogos-32 Dataset}
\label{sec:flickr32dataset}
\begin{figure*}[htbpp]
	\centering
	\includegraphics[width=0.85\textwidth]{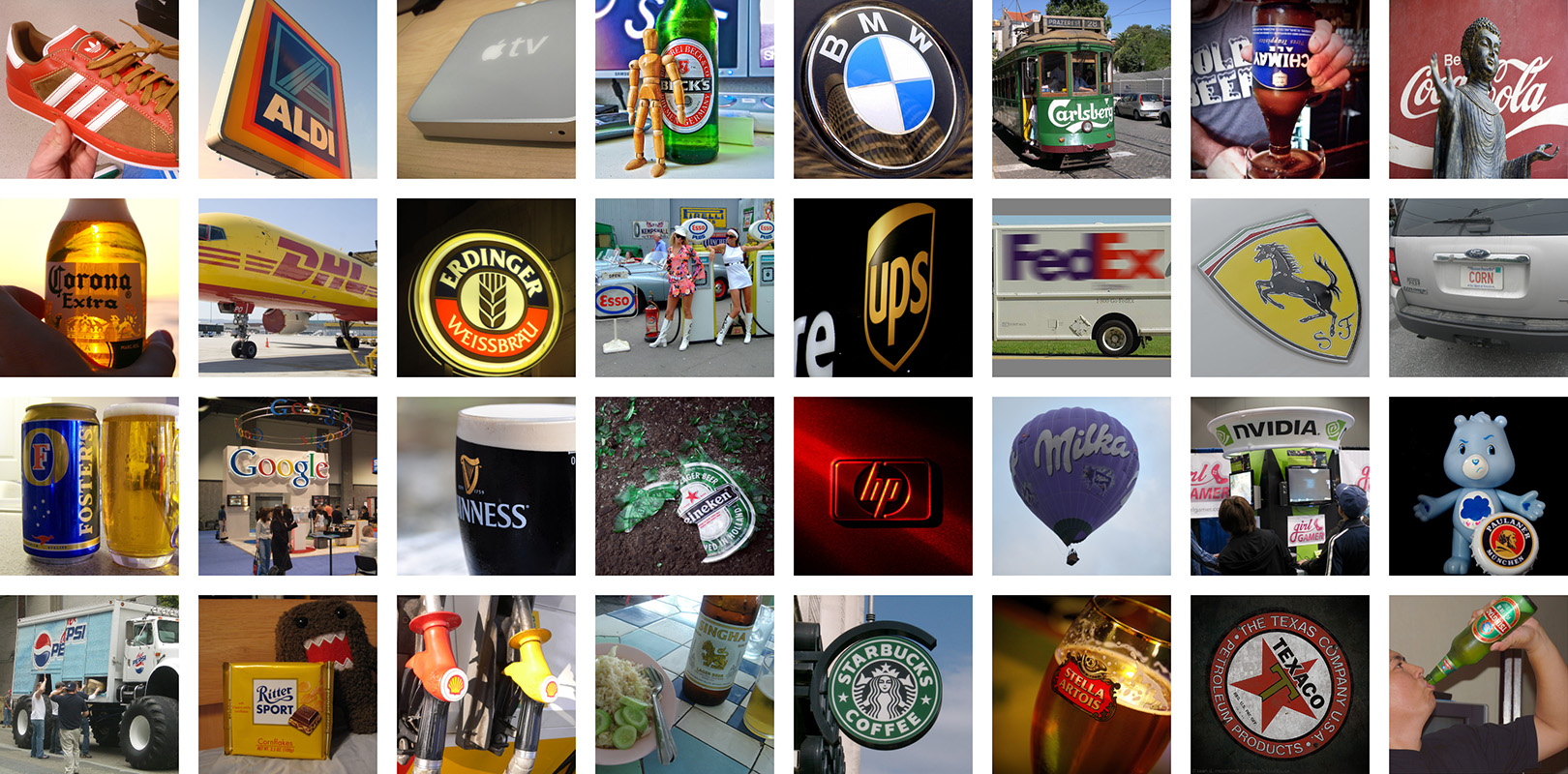}
	\caption{Example images for each of the 32 classes of the FlickrLogos-32 dataset.}
	\label{fig:example}
\end{figure*}

FlickrLogos-32 dataset \cite{romberg2011scalable} is a publicly-available collection of photos showing 32 different logo brands. It is meant for the evaluation of logo retrieval and multi-class logo detection/recognition systems on real-world images. All logos have an approximately planar or cylindrical surface.
For each class, the dataset offers 10 training images, 30 validation images, and 30 test images. An example image for each of the 32 classes of the FlickrLogos-32 dataset is reported in Figure~\ref{fig:example}.

%%%%%%%%%%%%%%%%%%%%%%%%%%%%%%%%%%%%%%%%%%%%%%%%%%%%%%%%%%%%%%%
%\section{IVL-Logos Dataset}
\subsection{Logos-32plus Dataset}
\label{sec:ivldataset}
Logos-32plus dataset is an expansion of the trainset of FlickrLogos-32. It has the same classes of objects as its counterpart but a larger cardinality (12312 instances).
We collected this new dataset for three main reasons: first, since we want to test a deep learning approach,
%instead of a keypoint-based one,
we needed a suitable dataset size. Second, we believe that Logos-32 dataset is not very representative of a data distribution for most real-world problems. Third, we hypothesize that synthetic data augmentation is not enough to model actual logo appearance variability.
The Logos-32 dataset was collected with the aim to train keypoint-based approaches. Therefore the selection of images followed some implicit guidelines, such as: most of the images are on focus, no blurry or noisy images, and usually images with highly saturated colors. As a result, the variability of this dataset mainly resides on the amount of intraclass affine transformations which can be handled very well by keypoint-based detection methods. We collected this new dataset with the aim of taking into account a larger set of real imaging conditions and transformations that may occur in uncontrolled acquisitions.

\begin{figure*}[tb]
	\includegraphics[width=\textwidth]{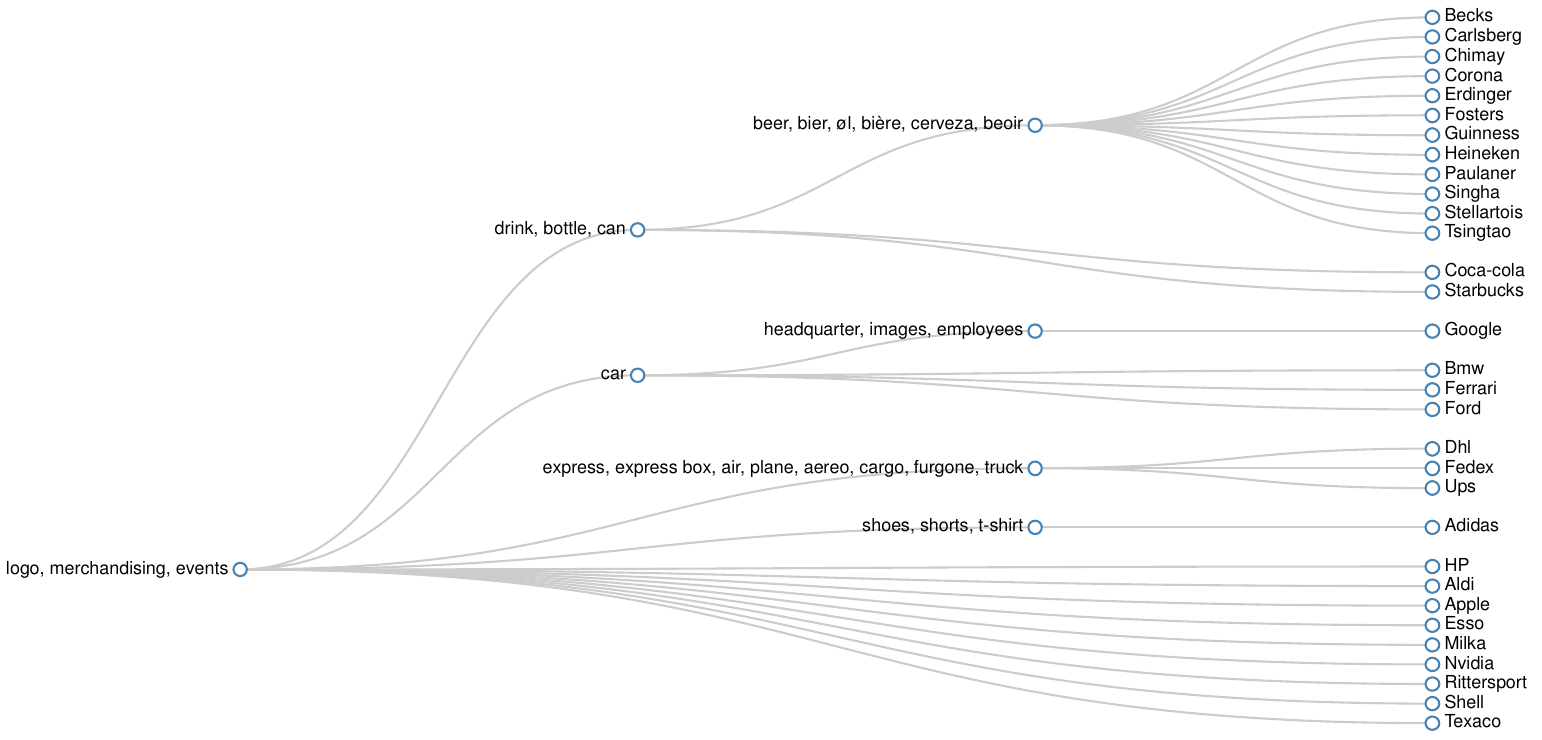}
	\centering
	\caption{Dendrogram representing the queries composition used to download the Logos-32plus dataset. To retrieve images of becks logos we used for instance: ``logo Becks'', ``merchandising Becks'', ``drink Becks'', ``bottle Becks'', ``beer Becks'' etc.}
	\label{fig:logostags_dendrogram}
\end{figure*}

%\subsection{Dataset construction}
%\subsubsection{Dataset construction}
%\label{sec:ivldataset_build}
We built the Logos-32plus dataset with images retrieved from both Flickr and Google image search. In particular, to increase the variability of data distribution we performed multiple queries for each logo. The dendrogram scheme in Figure \ref{fig:logostags_dendrogram} shows the tags used to compose the search queries used. To compose a single query we concatenate one leaf (a single logo) with a single tag of an ancestor node. The whole set of queries for each logo can be obtained by concatenating the logo name (leaf) with each tag contained in all the ancestors nodes. For example, all the queries used to search for the ``Becks'' logo are: ``logo Becks'', ``merchandising Becks'', ``events Becks'', ``drink Becks'', ``bottle Becks'', ``can Becks'', ``beer Becks'', ``bier Becks'' etc.

\begin{figure*}[htbp]
\centering
	\includegraphics[width=0.9\textwidth]{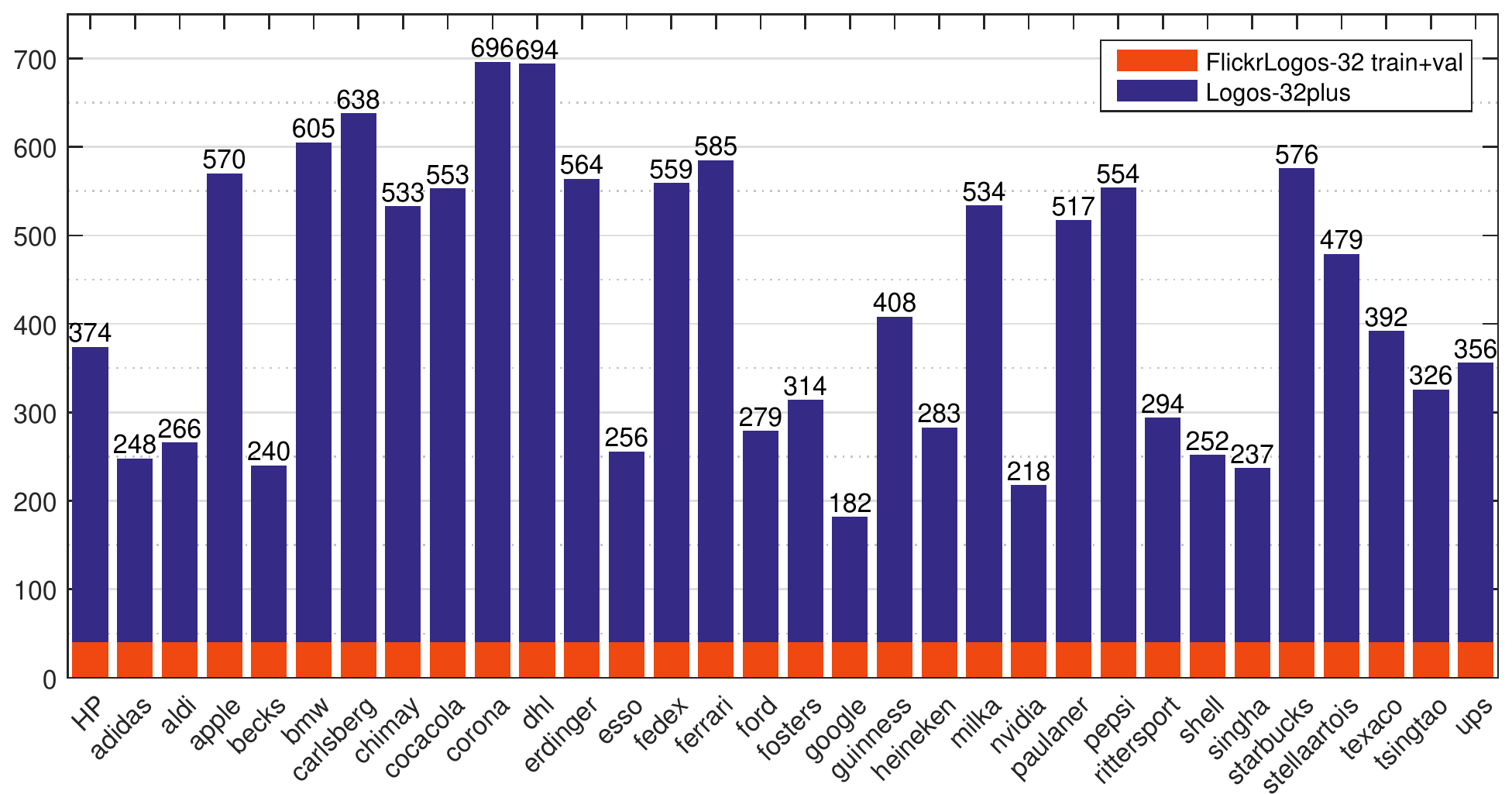}
	\caption{Graphical comparison of the distribution of the 32 logo classes between FlickLogos-32 and our augmented Logos-32plus dataset}
	\label{fig:dataset_histogram}
\end{figure*}

The dataset contains on average 400 examples per class, with each image including one or multiple instances of the same class. The detailed distribution of classes is shown in Figure \ref{fig:dataset_histogram} and a comparison between the FlickrLogos-32 and the Logos-32plus datasets is presented in Table \ref{tab:datasets_comparison}.
{The dataset is made available for research purposes at http://www.ivl.disco.unimib.it/activities/logo-recognition.}

%To ensure a high variability of the new dataset and to avoid any overlap with the existing one, we performed a check for duplicate images within the Logos-32plus dataset itself and with the FlickrLogos-32 dataset. The process has been carried out in two steps.
%First, we automatically found and discarded exact image duplicates using the SSIM measure \cite{wang2004image} \textcolor{red}{soglia?}.
%As a second step, we removed near duplicated images. To accomplish this task we trained a CNN from scratch (structure in Table \ref{tab:net_architecture}) on our collected logos. We truncated the network leaving out the last two layers (softmax and last fully-connected) and we used this network to extract features from every image in the two datasets. Finally we used a $k$-NN classifier to retrieve from Logos-32plus the examples most similar to FlickrLogos-32 test set, and checked manually for duplicates among the 5 nearest results.

%After near duplicate removal, the Logos-32plus dataset contains on average 400 examples per class, with each image including one or multiple instances of the same class. The detailed distribution of classes is shown in Figure \ref{fig:dataset_histogram} and a comparison between the FlickrLogos-32 and the Logos-32plus datasets is presented in Table \ref{tab:datasets_comparison}.

%{\color{black!40!green}
%The dataset will be made available at http://www.ivl.disco.unimib.it/ (link available after acceptance).%}

\subsection{Duplicates Removal}
\label{sec:duplicatesremoval}
To ensure a high variability of the new dataset and to avoid any overlap with the existing one, we performed a semi-automatic check for duplicate images within the Logos-32plus dataset itself and with the FlickrLogos-32 dataset. The process has been carried out in two steps. First, we automatically found and discarded image duplicates using the SSIM measure \cite{wang2004image}: we checked for similarity every pair of images within the Logos-32plus dataset itself and with the FlickrLogos-32 dataset using the SSIM measure. Images with SSIM measure over 0.9 have been discarded.

\begin{figure}[!h]
\centering
\begin{tabular}{c}
	\includegraphics[width=0.4\textwidth]{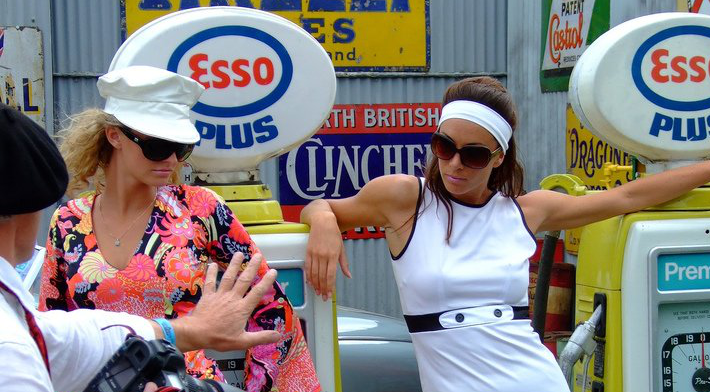} \\
	\includegraphics[width=0.4\textwidth]{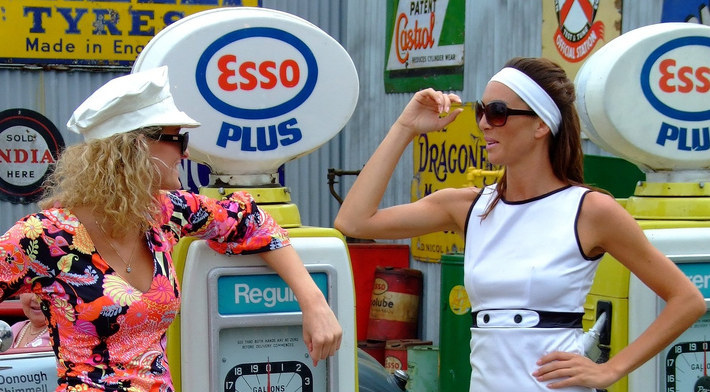}\\
\end{tabular}

	\caption{Example of near duplicates. The two images depict the same scene from a similar point of view. The appearance of the Esso logo in the two images is basically the same. We removed one of the two images from our Logos-32 plus dataset because the other one is included in the FlickrLogos-32 test set.}
	\label{fig:nearduplicates}
\end{figure}

As a second step, we removed near duplicates in a semi-automatic manner. % The definition of near duplicate is introduced in the next paragraph.\\
%\subsection{Exact Duplicates}
%\label{sec:duplicatesremoval}
%\subsection{Near duplicates}
%\label{sec:duplicatesremoval}
We say that two images are near duplicates if they depict the same scene with small differences in appearance with a particular focus on the portion of the image containing the logo. Examples of near duplicates are different overlapping crops of the same photo or images of the same scene from a different point of view. An interesting example of near duplicates is shown in Figure \ref{fig:nearduplicates}. The two images depict the same gas station from a very similar point of view. The girls in the photo are in different poses but the appearance of the Esso logo in the two images is basically the same. %We removed the right image from our FlickrLogos-32 dataset because the left image comes from FlickrLogos-32 test set.\\
In detail, to remove near duplicates we used the following procedure:
\begin{itemize}
\item[-] we trained our CNN (structure in Table \ref{tab:net_architecture}) from scratch on Logos-32plus dataset. To accomplish this task we fed the network with crops extracted from GT annotations and Object-proposals regions.
\item[-] We truncated the learned network leaving out the last two layers (softmax and last fully-connected). This network surgery operation let us use our network as a feature extractor exploiting the robust features learned by a deep neural network. We used this truncated network to extract features from every image crop that contains a tagged logo.
\item[-] We trained a k-NN classifier on top of the extracted features (using Logos-32plus as training set) and used it to retrieve from Logos-32plus and FlickrLogos-32 the nearest five results.
\item[-] Finally we manually checked for near duplicates among the five nearest results retrieved by the classifier. All the near duplicates have been discarded from the final dataset.
\end{itemize}

\begin{table*}[htbp]
	\caption{Comparison between FlickrLogos-32 and Logos-32plus datasets}
	\label{tab:datasets_comparison}
	\center
\begin{tabular}{lrr}
\toprule
 & \textbf{FlicrkLogos-32} & \textbf{Logos-32plus} \\ \midrule
Total images & 8240 & 7830\\
Images containing logo instances & 2240 & 7830\\
%Test annotations & 1602 & 0 \\
Train + Validation annotations & 1803 & 12302 \\
Average annotations for class (Train + Validation) & 40 & 400\\
Total annotations & 3405 & 12302 \\ \bottomrule
\end{tabular}
\end{table*}

%\begin{figure}[htbp]
%	\includegraphics[width=\textwidth]{augmentation.pdf}
%	\centering
%	\caption{Different effects of data augmentation. The first is real data augmentation, obtained by expanding FlickrLogos-32 with our Logos-32 plus dataset. The second is synthetic augmentation, obtained by translating the cropped region around the ground truth locations and around the object proposal locations.}
%	\label{fig:data_aug}
%\end{figure}

%%%%%%%%%%%%%%%%%%%%%%%%%%%%%%%%%%%%%%%%%%%%%%%%%%%%%%%%%%%%%%%
\section{Experimental Setup and Results}
\label{sec:results}

Experiments are performed considering the different training choices described in Section~\ref{sec:proposedmethod}. These include class balancing, data augmentation, image contrast normalization, sample weighting, addition of a background class, and addition of positive examples actually generated by the object proposal algorithm.\\
%First we evaluate the impact of including background negative examples in the training data (Training Configuration I and II).\\
%We then implement both real data augmentation using object proposals (TC III) and synthetic data augmentation using image translations (TC IV).\\
%Next we consider class balancing at two different levels: in TC V classes are balanced in each epoch only, while in TC VI classes are balanced in each batch as well. All subsequent tests are performed on both alternatives.\\
%The final experiments include contrast normalization as pre-processing (TC VII and IX) and sample weighting based on overlap between object proposal and ground-truth (TC VIII and X).
Each change to the training procedure is introduced one at a time, in order to assess its individual contribution, and the corrisponding value is underlined in Table \ref{tab:results_thval} for better readability.
All these configurations are trained using real data augmentation, i.e. with our extended Logos-32plus dataset in addition to FlickrLogos-32 training and validation sets.
Results are reported in Table \ref{tab:results_thval} in terms of both F1-measure and Accuracy on FlickrLogos-32 test set. With reference to Figure~\ref{fig:pipe_test}, the threshold on CNN predictions is automatically chosen to maximize the accuracy on FlickrLogos-32 training and validation sets.
The best configuration is then compared to other state of the art methods in Table~\ref{tab:comparison}.
As further investigation we quantify the contribution given from real data augmentation, by training the same solution on the original FlickrLogos-32 training set only.
Finally, we assess the impact of the object proposal algorithm to the overall performance. To do this we add all the ground truth locations to the test set, instead of relying on the object proposal only. %The difference in performance with respect to the proposed approach will provide an indication on how much of the final results depend on the quality of the object proposal instead of the aon the classification itself. %In fact in the latter case, among the proposal there are the logos correctly segmented.

\begin{table*}[!hhtbp]
	\caption{Experimental results showing the impact of the different training choices described in Section~\ref{sec:proposedmethod} on the final classification. Results are reported in terms of Precision, Recall, F1-measure and Accuracy.}
	\label{tab:results_thval}
	%\center

\begin{tabular}{c}

\begin{tabularx}{1\textwidth}{@{}XXXXXXX|XXXX@{}}
\toprule
\textbf{Train. Config.} & \textbf{BG class} & \textbf{BBs}    & \textbf{Data Augm.}    & \textbf{Class bal.} & \textbf{Contr. norm.}  & \textbf{Sample weight} & \textbf{Prec.} & \textbf{Rec.} & \textbf{F1} & \textbf{Acc.} \\ \midrule
I & No              & GT                & No       & No         & No        & No     & 0.370 & 0.370 & 0.370 & 0.096 \\
II & \underline{Yes} & GT                & No       & No         & No        & No     & 0.713 & 0.665 & 0.688 & 0.620 \\
III & Yes             & \underline{GT+OP} & No       & No         & No        & No     & 0.816 & 0.787 & 0.801 & 0.744 \\
IV & Yes             & GT+OP             & \underline{Yes}      & No         & No        & No     & 0.987 & 0.858 & 0.918 & 0.953 \\
V & Yes             & GT+OP             & Yes      & \underline{Epoch}    & No        & No     & 0.986 & 0.865 & 0.922 & 0.956 \\
VI & Yes             & GT+OP             & Yes      & \underline{Batch}    & No        & No     & 0.980 & 0.833 & 0.901 & 0.945 \\
VII & \textbf{Yes}             & \textbf{GT+OP}             & \textbf{Yes}      & \textbf{\underline{Epoch}}    & \textbf{\underline{Yes}}       & \textbf{No}     & \textbf{0.989} & \textbf{0.906} & \textbf{0.946} & \textbf{0.958} \\
VIII & Yes             & GT+OP             & Yes      & Epoch    & Yes       & \underline{Yes}    & 0.984 & 0.875 & 0.926 & 0.951 \\
IX & Yes             & GT+OP             & Yes      & \underline{Batch}    & \underline{Yes}       & No     & 0.984 & 0.887 & 0.933 & 0.955 \\
X & Yes             & GT+OP             & Yes      & Batch    & Yes       & \underline{Yes}    & \textbf{0.989} & 0.866 & 0.923 & 0.955 \\ \bottomrule
\end{tabularx}
%\end{table}

\\
\\

\footnotesize
%\begin{table}[]
\centering
%\caption{My caption}
%\label{my-label}
\begin{tabular}{@{}lll@{}}
\toprule
\multicolumn{3}{c}{\textbf{Legend to Table \ref{tab:results_thval}}}                                                                       \\ \midrule
\textbf{Train. Config.}   & \multicolumn{2}{l}{Identifier of the configuration used for training} \\
\textbf{BG class}   & \multicolumn{2}{l}{Background class (no-logo examples) included in training}        \\
\textbf{BBs}        & \multicolumn{2}{l}{Bounding Boxes used as training examples}                        \\
                    & \textbf{GT}                 & Precise ground-truth logo annotations                                     \\
                    & \textbf{GT+OP}              & Precise ground-truth and Object-proposal logo annotations           \\
\textbf{Data Augm.} & \multicolumn{2}{l}{Data Augmentation (translation)}                                 \\
\textbf{Class bal.} & \multicolumn{2}{l}{Class balancing to account for different cardinalities}         \\
                    & \textbf{Epoch}            & Classes are balanced in each epoch                    \\
                    & \textbf{Batch}            & Classes are balanced in each batch as well            \\
\textbf{Contr. norm.}  & \multicolumn{2}{l}{Pre-processing of training examples with contrast normalization} \\
\textbf{Sample weight}     & \multicolumn{2}{l}{Weighting examples based on overlap between OP and GT}           \\ \bottomrule
\end{tabular}
%\end{table}%

\end{tabular}
\end{table*}

%{\color{black!40!green}
From the results reported in Table~\ref{tab:results_thval} it is possible to see that with respect to a straightforward application of deep learning to the logo recognition task (i.e. Training Configuration I, \emph{TC}-I), the different training choices considered are able to give a large increase in performance:
\begin{itemize}
\item[-] The first jump in performance is obtained by including the background (i.e. no-logo examples) as a new class in training. Results are identified as \emph{TC}-II and show an improvement in F1-measure and accuracy of 31.8\% and 52.4\% with respect to \emph{TC}-I.
\item[-] A second jump is obtained by including object proposals coming from Selective Search as additional training examples. This configuration is named \emph{TC}-III and improves the F1-measure and accuracy by 11.3\% and 12.4\% with respect to \emph{TC}-II.
\item[-] A third jump in performance is obtained by augmenting the cardinality of object proposals coming from Selective Search by perturbing them with random translations (i.e. synthetic data augmentation). This configuration is named \emph{TC}-IV and improves the F1-measure and accuracy by 11.7\% and 20.9\% with respect to \emph{TC}-III.
\item[-] A further, smaller, improvement in performance is obtained by considering class balancing to account for different cardinalities, with ``Epoch'' balancing giving consistently better performance than the ``Batch'' counterpart (named \emph{TC}-V and \emph{TC}-VI respectively). In particular, \emph{TC}-V improves the F1-measure and accuracy by 0.4\% and 0.3\% with respect to \emph{TC}-IV.
\item[-] Contrast normalization brings a further little but consistent improvement, with \emph{TC}-VII improving the F1-measure and accuracy by 2.4\% and 0.2\% with respect to \emph{TC}-V.
\item[-] Sample weighting instead (adopted in \emph{TC}-VIII and \emph{TC}-X), which consists in weighting training examples according to the degree of overlap between the object proposal and ground truth regions, results in lowering the final performance of the method.
\end{itemize}
The best configuration (i.e. \emph{TC}-VII) trained on our extended training set is highlighted in bold in Table~\ref{tab:results_thval} and compared with the state of the art in Table~\ref{tab:comparison}. Performances of the other methods are taken from the respective papers and thus for some of them some performance measures are missing.
From the results reported it is possible to see that the proposed solution is able to improve the F1-measure with respect to the best method in the state of the art by 3.8\%, and the accuracy by 1.7\%. It is worth to underline that the best results for the two metrics were obtained by different methods in the state of the art, i.e. by  Romberg et. al \cite{romberg2013bundle} and BoW SIFT \cite{romberg2013bundle} respectively.

\begin{figure*}[htbp]
\setlength{\tabcolsep}{6pt}
{\scriptsize
\resizebox{\textwidth}{!}{
\begin{tabular}{p{0.75cm}p{0.75cm}p{0.75cm}p{0.75cm}p{0.75cm}p{0.75cm}p{0.75cm}p{0.75cm}p{0.75cm}p{0.75cm}p{0.75cm}p{0.75cm}p{0.75cm}p{0.75cm}p{0.75cm}}
Chimay & Adidas & Ritter & HP & Ritter & CocaCola & Ritter & BMW & Apple & CocaCola & Adidas & Erdinger & Ritter & Erdinger & FedEx\\
\end{tabular}}
}
	\includegraphics[width=\textwidth]{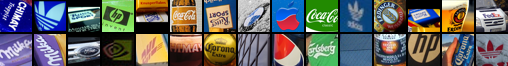}
	\vskip -0.15truecm
\setlength{\tabcolsep}{6pt}
{\scriptsize
\resizebox{\textwidth}{!}{
\begin{tabular}{p{0.75cm}p{0.75cm}p{0.75cm}p{0.75cm}p{0.75cm}p{0.75cm}p{0.75cm}p{0.75cm}p{0.75cm}p{0.75cm}p{0.75cm}p{0.75cm}p{0.75cm}p{0.75cm}p{0.75cm}}
Milka & Milka & Ford & Nvidia & DHL & Chimay & Corona & Backgr. & Pepsi & Carlsberg & Backgr. & Corona & HP & Corona & Adidas\\
\end{tabular}}
}

	\vspace{0.25truecm}

\setlength{\tabcolsep}{6pt}
{\scriptsize
\resizebox{\textwidth}{!}{
\begin{tabular}{p{0.75cm}p{0.75cm}p{0.75cm}p{0.75cm}p{0.75cm}p{0.75cm}p{0.75cm}p{0.75cm}p{0.75cm}p{0.75cm}p{0.75cm}p{0.75cm}p{0.75cm}p{0.75cm}p{0.75cm}}
Aldi & Adidas & Nvidia & Apple & Erdinger & Shell & Tsingtao & Carlesberg & Adidas & Nvidia & Nvidia & Ferrari & Tsingtao & Apple & UPS\\
\end{tabular}}
}
	\includegraphics[width=\textwidth]{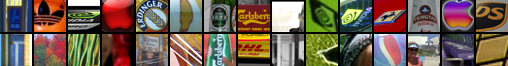}
%	\includegraphics[width=\textwidth]{pipe_training.png}
%	\centering
\vskip -0.15truecm
\setlength{\tabcolsep}{6pt}
{\scriptsize
\resizebox{\textwidth}{!}{
\begin{tabular}{p{0.75cm}p{0.75cm}p{0.75cm}p{0.75cm}p{0.75cm}p{0.75cm}p{0.75cm}p{0.75cm}p{0.75cm}p{0.75cm}p{0.75cm}p{0.75cm}p{0.75cm}p{0.75cm}p{0.75cm}}
Backgr. & Backgr. & Backgr. & Backgr. & Corona & Backgr. & Carlsberg & DHL & Backgr. & Backgr. & Backgr. & Pepsi & Backgr. & Milka & Backgr.\\
\end{tabular}}
}
	\caption{Wrongly labeled logos ordered by confidence. Highest confidence prediction is top-left. Images resolution is 32x32 pixels, i.e. the same used to feed the CNN. The first and third rows are the wrong labeled logos, the second and the fourth rows represent the nearest example in the training set (using the last network layer activations before the softmax as feature vector).}
	\label{fig:errors}
\end{figure*}

As a further comparison, we report the results obtained by our solution using only FlickrLogos-32 for training and keeping all the other training choices unchanged. This results in a drop in F1-measure by 14.7\% and by 4.8\% in accuracy, giving an idea of the benefit of real data augmentation with respect to a purely synthetic one \cite{eggert2015benefit}.
%Since in the preliminary verision of this paper \cite{bianco2015logo} we showed that Selective Search As a final analysis
As a final analysis, to understand if the major source of error in our method is the Selective Search module that is unable to have a high recall or if its the CNN itself that mispredicts the logo class, we perform an additional test by adding the actual logo ground truth region to the object proposals. This increases the F1-measure by 0.6\% and the accuracy by 0.2\% indicating that its the major source of error in our method is the CNN itself. Some examples of wrongly labeled candidate logo regions are reported in Figure \ref{fig:errors}. Candidates are generated by the object proposal and they have a IoU larger than 0.5 with the corresponding ground truth.  The first and the third row depict the wrongly recognized regions labeled with their actual class, while the second and fourth one depict the nearest example in the training set using as features the activations of the last network layer before the softmax. Images are reported with the same resolution used to feed the CNN, i.e. 32$\times$32 pixels.
%}

%\subsection{Comparison with other methods}

\begin{table*}[htbp]
	\caption{Comparison of the best configuration in Table \ref{tab:results_thval} with the methods in the state of the art.}
	\label{tab:comparison}
	\center
\begin{tabularx}{0.97\textwidth}{ll|rrrr}
\toprule
\textbf{Method} & \textbf{Train data} & \textbf{Precision}               & \textbf{Recall}   & \textbf{F1}                 & \textbf{Accuracy} \\ \midrule
BoW SIFT \cite{romberg2013bundle}   & FL32               & 0.991              & 0.784 & 0.875              & 0.941 \\
BoW SIFT + SP + SynQE \cite{romberg2013bundle} & FL32& 0.994              & 0.826 & 0.902              & N/A   \\
Romberg et al. \cite{romberg2011scalable}   & FL32          & 0.981              & 0.610 & 0.752              & N/A   \\
Revaud et al. \cite{revaud2012correlation}  & FL32           & $\geq$0.980 & 0.726 & 0.834$\div$0.841 & N/A   \\
Romberg et al. \cite{romberg2013bundle}   & FL32          & \textbf{0.999}              & 0.832 & 0.908              & N/A   \\
Bianco et al. \cite{bianco2015logo}   & FL32          & 0.909              & 0.845 & 0.876              & 0.884   \\
Bianco et al. + Q.Exp. \cite{bianco2015logo}  & FL32           & 0.971              & 0.629 & 0.763              & 0.904   \\
Eggert et al. \cite{eggert2015benefit}  & FL32         & 0.996              & 0.786 & 0.879              & 0.846   \\
{Oliveira et al. \cite{oliveira2016automatic}}   & {FL32}          & {0.955 }               & {\textbf{0.908}}   & {0.931}                & {N/A}  \\
DeepLogo \cite{iandola2015deeplogo}   & FL32          & N/A                & N/A   & N/A                & 0.896 \\ \midrule
Ours (\emph{TC}-VII) & FL32                 & 0.976              & 0.676           & 0.799              & 0.910 \\
Ours (\emph{TC}-VII) & FL32, L32+            & 0.989              & 0.906  & \textbf{0.946}     & \textbf{0.958} \\ \midrule
Ours (\emph{TC}-VII, adding GT to the obj. prop.) & FL32            & 0.968              & 0.755           & 0.848              & 0.917 \\
Ours (\emph{TC}-VII, adding GT to the obj. prop.) & FL32, L32+        & 0.989              & \textbf{0.917}  & \textbf{0.952}     & \textbf{0.960} \\ \bottomrule
\end{tabularx}
\end{table*}

\subsection{Timings}

\begin{table}[htbp]
	\caption{Timings of the whole recognition pipeline}
	\label{tab:timings}
	\centering
\resizebox{\columnwidth}{!}{
\begin{tabular}{l|l l l l}
\toprule
\textbf{Device} & \textbf{Proposal} & \textbf{Preproc.} & \textbf{Classif.} & \textbf{Overall}\\ \midrule
CPU & 1.24 s & 0.93 s & 0.71 s & 2.91 s \\
GPU & 1.24 s & 2.12 s & 0.36 s & 3.74 s \\ \bottomrule
\end{tabular}}
\end{table}
Table \ref{tab:timings} shows the timings for the whole recognition procedure at test time. Experiments are performed on the same computer (Intel i7 3.40 GHz - 16 GB RAM) averaging the timings of 100 runs on different images.\\
Two different solutions are compared: the use of CPU or GPU (GeForce GTX 650) for the classification step.\\
The proposals extraction step runs always on CPU. The prepocessing time include the resize of every patch to match the CNN input size, the contrast normalization (negligible processing time) and eventually the time to copy the data from CPU to GPU memory. In Table \ref{tab:timings} it is possible to notice that the overhead caused by the CPU-GPU memory transfer makes the overall time of the GPU solution higher than that of the CPU solution. {To this extent, in the future it might be interesting to evaluate a fully GPU-based pipeline, for example generating and pre-processing proposals according to \cite{ren2015faster}.}

%%%%%%%%%%%%%%%%%%%%%%%%%%%%%%%%%%%%%%%%%%%%%%%%%%%%%%%%%%%%%%%
\section{Conclusions}
\label{sec:conclusions}
%{\color{black!40!orange}
Logo recognition is fundamental in many application domains. The problem is that logos may appear in any position, scale and under any point of view in an image. Moreover, the images may be corrupted by many image artifacts and distortions. %not have in contextual ad placement, validation of product placement, and online brand management.

%In this work we treated the problem of logo recognition.
The traditional approaches to logo recognition involve keypoint-based detectors and descriptors, or the use of  CNNs pretrained on different tasks. %however in the last year several  works started investigating the use of deep learning to address this problem.
%Given the limited size of publicly available datasets, none of the state of the art methods could train a
%network from scratch, using instead networks that were pretrained on a different task.\\
Our solution employs a %is composed of a candidate logo region proposal, followed by a
CNN specifically trained for the task of logo classification, even if they are not perfectly localized.
We designed a complete recognition pipeline including a recall-oriented candidate logo region proposal that feeds our CNN.

Experiments are carried out on the FlickrLogos-32 database and on its enlarged version, Logos-32plus, collected by the authors. We systematically investigated the effect on recognition performance of synthetic versus real data augmentation, image pre-processing, and the benefits of different training choices such as class-balancing, sample-weighting and explicit modeling the background class (i.e. no-logo regions).
Our best solution outperforms the methods in the state of the art and makes use of an explicit modeling of the background class, both precise and actual object-proposal logo annotations during training, synthetic data augmentation, epoch-based class balancing, and image contrast normalization as pre-processing, while sample weighting is disabled. Both the newly collected Logos-32plus and the trained CNN are made available for research purposes\footnote{http://www.ivl.disco.unimib.it/activities/logo-recognition}.

\bibliographystyle{elsarticle/elsarticle-num}
\bibliography{bibliography}


%\begin{thebibliography}{00}
%%% Text of bibliographic item
%
%\bibitem{}
%
%\end{thebibliography}
\end{document}